\documentclass[runningheads]{llncs}
\usepackage{graphicx}
\usepackage{amsmath}
\usepackage{booktabs} 
\usepackage{pifont}
\newcommand{\cmark}{\ding{51}}%
\newcommand{\xmark}{\ding{55}}%
\usepackage{xcolor}
\usepackage{float}
\usepackage[colorlinks = true,
            linkcolor = blue,
            urlcolor  = blue,
            citecolor = blue,
            anchorcolor = blue]{hyperref}
\usepackage{array}
\newcolumntype{P}[1]{>{\centering\arraybackslash}p{#1}}
\newcommand{\DFLB}{{\mathchoice{}{}{\scriptscriptstyle}{}DFLB}}
\newcommand{\CGL}{{\mathchoice{}{}{\scriptscriptstyle}{}CGL}}
\newcommand{\CE}{{\mathchoice{}{}{\scriptscriptstyle}{}CE}}
\newcommand*\xor{\oplus}
\newcommand{\MYhref}[3][blue]{\href{#2}{\color{#1}{#3}}}

\begin{document}
\title{Catch Me if You Can: A Novel Task for Detection of Covert Geo-Locations (CGL)}
%
%
\author{Binoy Saha\inst{1}\orcidID{0000-0002-8402-2412} \and
Sukhendu Das\inst{2}\orcidID{0000-0002-2823-9211}}
\authorrunning{S. Binoy et al.}
\titlerunning{Catch Me if You Can: A Novel Task for CGL Detection}
%
\institute{
\email{binoysaha@cse.iitm.ac.in}\\ \and
\email{sdas@iitm.ac.in}\\
Visualization and Perception Lab, Dept. of CSE, IIT Madras, India}
%
\maketitle              
\begin{abstract}
Most of the visual scene understanding tasks in the field of computer vision involve identification of the objects present in the scene. Image regions like hideouts, turns, and other obscured regions of the scene also contain crucial information, for specific surveillance tasks. In this work, we propose an intelligent visual aid for identification of such locations in an image, which has either the potential to create an imminent threat from an adversary or appear as the target zones needing further investigation to identify concealed objects. Covert places (CGL) for hiding behind an occluding object are concealed 3D locations, not usually detectable from the viewpoint (camera). Hence this involves delineating specific image regions around the outer boundary of the projections of the occluding objects, as places to be accessed around the potential hideouts.  CGL detection finds applications in military counter-insurgency operations, surveillance with path planning for an exploratory robot. Given an RGB image, the goal is to identify all CGLs in the 2D scene. Identification of such regions would require knowledge about the 3D boundaries of obscuring items (pillars, furniture), their spatial location with respect to the background, and other neighboring regions of the scene. We propose this as a novel task, termed Covert Geo-Location (CGL) Detection. Classification of any region of an image as a CGL (as boundary sub-segments of an occluding object, concealing the hideout) requires examining the relation with its surroundings. CGL detection would thus require understanding of the 3D spatial relationships between boundaries of occluding objects and their neighborhoods. Our method successfully extracts relevant depth features from a single RGB image and quantitatively yields significant improvement over existing object detection and segmentation models adapted and trained for CGL detection. We also introduce a novel hand-annotated CGL detection dataset containing ~1.5K real-world images for experimentation.\\
\textbf{ACK: IMPRINT (MHRD/DRDO) GoI, for support}
\end{abstract}
\section{Introduction}
\begin{figure}[!t]
  \centering
  \includegraphics[scale=0.45]{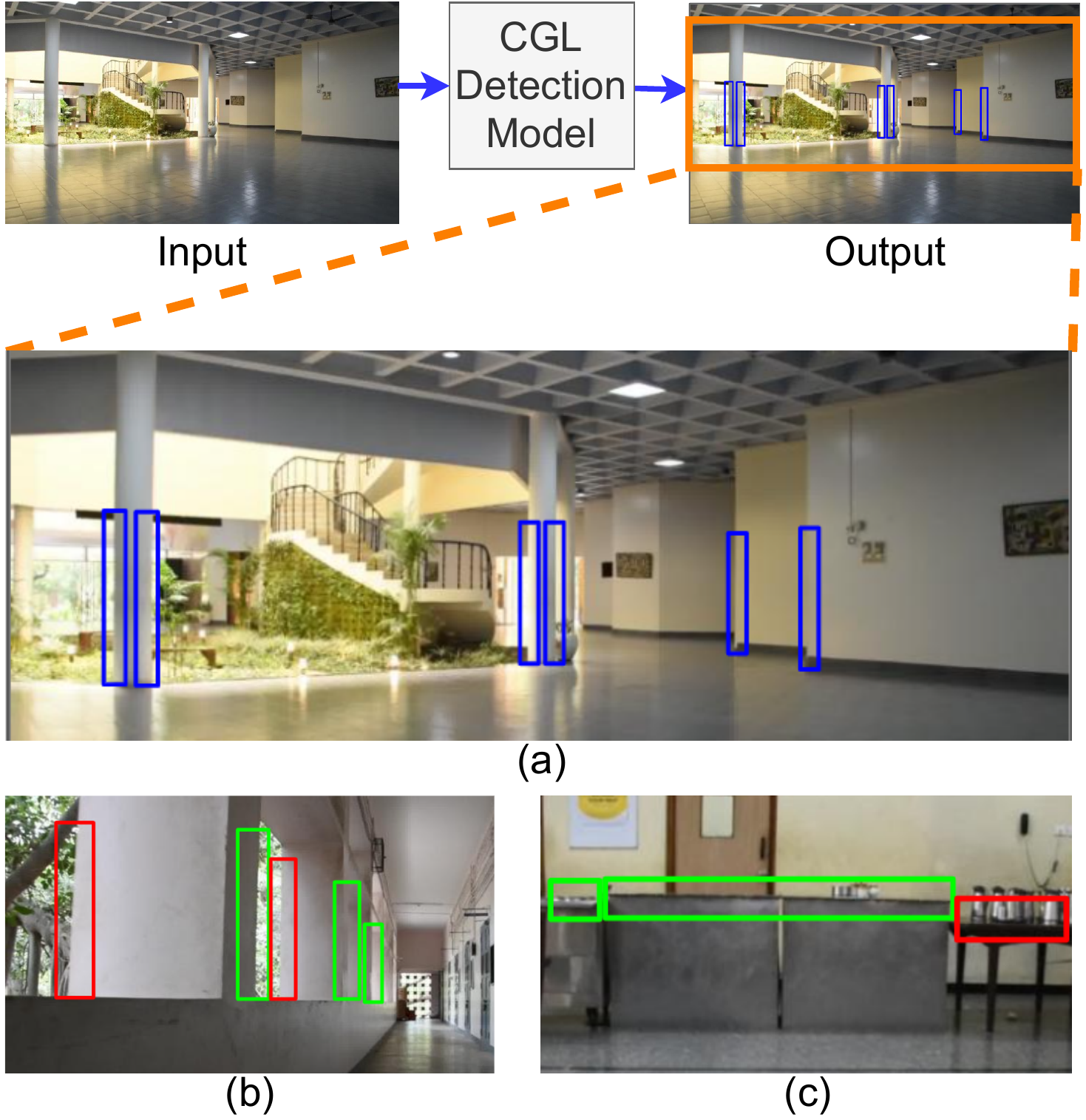}
  \caption{(a) shows the input and the expected output for CGL detection. Blue bounding boxes indicate hand-annotated CGL Ground Truth. In (b) and (c), green bounding boxes indicate valid CGL locations; while red ones indicate locations that appear similar to a CGL, but do not fall in that category (hideout or covert locations). In case of (b), no floor or platform exists at the base of the vertical zones (behind the pillar edges), indicated by red bounding boxes, for anything to be placed and be obscured. In case of (c), the obscuring item (tabletop) cannot hide anything behind it.
  }
  \label{intro_fig}
\end{figure}

The primary focus of the computer vision community has been on the understanding of visual scenes. To this end, many datasets and tasks have been proposed over the years to build AI systems that can perform specific scene understanding tasks like image classification \cite{pham2020meta,foret2020sharpness,kabir2020spinalnet,byerly2020branching,simonyan2014very,szegedy2015going,he2016deep}, object detection \cite{girshick2014rich,liu2016ssd,farhadi2018yolov3,lin2017feature,tan2020efficientdet}, image captioning
\cite{cornia2020meshed,li2020oscar,zhang2021vinvl,zhou2020unified}, visual question
answering \cite{VQA,balanced_binary_vqa,balanced_vqa_v2,li2020oscar,zhang2021vinvl}, scene graph generation \cite{li2017scene,yang2018graph,tang2019learning,tang2020unbiased} and many more. With the
advent of deep convolutional networks \cite{lecun1998gradient,krizhevsky2012imagenet,simonyan2014very,szegedy2015going}, one of the tasks where AI has evolved remarkably is object detection. Most successful object detection approaches are either based on two-stage region
proposal methods \cite{ren2015faster,he2017mask} or single-shot detection methods \cite{lin2017focal,zhao2019m2det}. These methods have become exceptionally good at recognizing and detecting
objects. Thus, object detection can be used to instill knowledge about
the distinguishing features of objects into machines.

Although, most of the visual scene understanding tasks in the field of computer vision involve identification of the objects present in the scene, non-object image regions like hideouts, corners, bends, turns, and other obscured regions of the scene also contain crucial information, for a specific set of surveillance tasks. Thus, the next step for advancement towards the goal of scene understanding would be to detect such covert locations in a scene.

Covert places for hiding behind any occluding objects (pillars, doors, furniture), are concealed locations that are not usually detectable from the viewpoint (camera). However, an intelligent agent can analyze the scene to foresee such potential regions around the obscuring (occluding) objects, from where the threats loom large. These specific parts (image regions) around the boundaries of occluding objects are the targets for detection by the algorithm. These locations may either be: (i) potential zones of threat caused by an adversary hiding behind the object, or (ii) target zones for further investigation to detect any other unidentified concealed object. To advance towards the goal of identification of such regions, we propose a novel task termed Covert Geo-Location (CGL) Detection. CGL detection finds applications in military counter-insurgency operations and intelligent scene surveillance (for identification of target zones) with path planning for a robot. In this work, we provide an intelligent visual aid for identification of such locations.

We define the problem addressed in this paper, as: Given an input RGB image, the goal is to identify all covert places (hideouts) in that image. We pose the CGL detection problem as that involving identification of regions encapsulating specific target boundary sub-segments of an obscuring item/object in a scene, which either poses threat or can act as target zones for further inspection of the scene. Since it is not possible to classify any region of an image as a hideout (CGL) without looking at its surroundings, CGL detection requires context-aware detection, and identification of CGLs in an image would require depth information of regions around the boundaries of obscuring items (like pillars, doors, furniture, wall endings, sofa, cot, etc.) and knowledge about spatial position (in 3D) of the occluding object with respect to its background and neighboring objects in the scene. 

We also highlight the importance of extracting depth information for CGL detection, which the traditional (object) detection approaches often do not consider. Our proposed method successfully extracts relevant depth features from only a single RGB image as input and quantitatively yields significant improvement over existing object detection and segmentation models (when adapted and trained for CGL detection).

In methods used for object detection, the classification of any region of interest (RoI) is solely dependent on the features of the RoI itself. Identifying and localizing covert locations will require additional knowledge about the RoI. For instance, to classify any object as an obstruction, the image region containing just the occluding object is not enough; knowledge about its position on the floor and the objects around it are equally important. As identification of covert locations requires knowledge about neighboring and background objects as well as their relative positions, models should learn to perform context-aware detection capturing the 3D spatial relationships between objects (their boundaries) and their surroundings.

For that purpose, we introduce a novel task termed Covert Geo-Location (CGL) Detection. Given an input RGB image, the goal is to identify and localize parts of boundaries of occluding objects, behind which lie potential hideouts (Covert Geo-Locations) in the scene. The output of the algorithms must hence be bounding box templates overlaying specific sections of boundaries of the occluding objects (as shown in figure \ref{intro_fig}.(a) ). This task is challenging because the model needs to infer the spatial relationship of an occluding object with the floor (for support, space requirements, and reachability), characteristics of boundaries on the obscuring items, etc. Figure \ref{intro_fig}.b and \ref{intro_fig}.c show examples of both cases: CGL (green boxes) as well as locations that appear similar to a CGL but actually are not (red boxes) so. In case of (b), no floor or platform exists at the base of the vertical zones (behind the pillar edges), indicated by red bounding boxes, for anything to be placed and be obscured. In case of (c), the obscuring item (tabletop) cannot hide anything behind it which rests on the floor.

We also present a novel dataset for CGL detection containing real-world images depicting diverse environments captured in normal lighting conditions. Our dataset consists of images captured in indoor environments and building premises. On an average, images contain around 8 to 9 CGLs, with the total number of hand-annotated CGLs being $\approx$15K. We believe that this first-of-a-kind dataset along with its analysis presented in this paper will offer a new domain of work along the line of the aforementioned application.

Humans identify hideouts by searching for areas in the scene where there is enough distance between the foreground and its background. Since a depth map captures the relative distance information, it can act as a good cue for CGL detection. As depth is independent of illumination, it provides a great complementary modality. However, effectively fusing information from RGB and depth modality is non-trivial, and it often requires additional computational power. Also, most of the existing datasets \cite{lin2014microsoft,Everingham15,krishnavisualgenome,balanced_vqa_v2} for vision-related tasks, including our CGL detection dataset, do not contain corresponding depth maps for input images. This leads us to the question, can we train models to jointly generate RGB and depth features, taking only RGB images as input? We address this question in this paper by proposing a novel method for CGL detection.

We observe that object detection models fail to capture the spatial contextual information essential for CGL detection. To perform context-aware detection, we thus map CGL detection to a binary segmentation task and experiment with segmentation models. We propose a novel segmentation-based method that implicitly attempts to train the feature extractor to generate relevant depth features. Our method uses an additional decoder called Depth-aware Feature Learning Block (DFLB), which is trained to classify image regions that have depth pattern similar to CGLs, as one class (say “potential CGL”), and the rest of the image as another class. DFLB serves two purposes, it aids the CGL segmentation block (supervised using hand-annotated ground truth) in attending to image regions having the necessary depth patterns appearing over CGL templates, and helps the common encoder to extract relevant depth features from the input RGB image. Subsequently, we propose two feature-level loss functions for the self-supervision of the common encoder, which further enhances the performance of our model.

\textbf{Our key contributions are summarized as follows:}
1)	We propose a novel CGL detection task, which requires context-aware location detection.
2)	We present a CGL detection dataset depicting diverse and challenging environments. Images from our dataset contain naturally occurring glares and shadows.
3)	We propose a method for jointly learning RGB and depth features taking only a single RGB image as input. We introduce the Depth-aware Feature Learning Block (DFLB), which steers the feature extractor towards extraction of relevant depth features.
4)	We propose two feature-level loss functions, namely Geometric Transformation Equivariance Loss (GTE loss) and Intraclass Variance Reduction loss (IVR loss) for self-supervision of the encoder.

\section{Related Work}
The ultimate goal of artificial intelligence is to mimic human intelligence. Throughout the history of artificial intelligence, the research community has tried to incrementally move towards this goal by designing tasks and curating datasets for incorporating various aspects of human intelligence into machines. In this work, we propose yet another task that can be used to instill a new type of knowledge into machines. In this section, we briefly describe some of the tasks that have been explored so far for visual scene understanding and contrast each one of them with our proposed task.

\textbf{Object Detection:}
In object detection \cite{ren2015faster,he2017mask,lin2017focal,zhao2019m2det}, the task is to identify and localize all objects present in the scene. For the classification of RoIs (regions that potentially contain an object) in object detection, features of the ROIs alone are sufficient. Global (image-level) context also influences the detection in certain cases but the local spatial context (knowledge about the surroundings of the RoI) is never exploited in object detection. On the other hand, local spatial context with respect to other neighboring regions of the scene is of paramount importance for CGL detection.

\textbf{Semantic Segmentation} 
The goal in semantic segmentation \cite{tao2020hierarchical,sun2019high,zoph2020rethinking} is to generate a pixel-level classification map having a size of $H \times W$, where each pixel is assigned to one of the K categories. Although CGL detection can be mapped to a binary semantic segmentation task, it is far more challenging to segment CGLs than the categories available in existing segmentation datasets \cite{zhou2017scene,zhou2018semantic,lin2014microsoft}.

\textbf{Scene Recognition:}
In this task, given an input image, the model needs to classify the entire image into one of the 'K' scene categories. As pointed out in \cite{zhou2015places2,zhou2017places,zhou2014learning}, for an intelligent system to understand the environment thoroughly, it should be able to identify the place, and to do that, the system needs to understand what set of objects co-occur at what places and how the arrangement of objects depends upon the place. As CGL detection encourages context-aware location detection, CGL detection models can aid in identifying complex regions like corners, hideouts, bends, turns and other obscuring regions of the scene, leading to better scene recognition.

\textbf{Other visual scene understanding tasks:}
Tasks like visual question answering VQA \cite{VQA,balanced_binary_vqa,balanced_vqa_v2} (which involves answering natural language questions about an input image), scene graph generation SGG \cite{krishnavisualgenome,tang2020unbiased,li2018factorizable} (which involves detection of objects in the input image and generation of a graph depicting the relationships between the detected objects), and image captioning \cite{lin2014microsoft,li2020oscar,cornia2020meshed} (which involves producing natural language captions for an input image) also require complex understanding of the relationships between the objects present in the scene, but RoI classification in all the above mentioned tasks is still context agnostic.

None of the datasets used for existing vision-related tasks has manually annotated regions in images identifying locations of hideout (regions for concealment) to be exploited for the problem addressed in this paper. CGLs can also be considered to be non-object locations (but in the vicinity of objects) around the edges of certain objects, having a distinct depth pattern. In surveillance scenarios, CGL detection can aid in the exploration of zones in an image, which are potential candidates for hazards and threats.

\section{Dataset Details}
\subsection{Images}
\begin{figure*}[ht]
  \centering
  \includegraphics[scale=0.5]{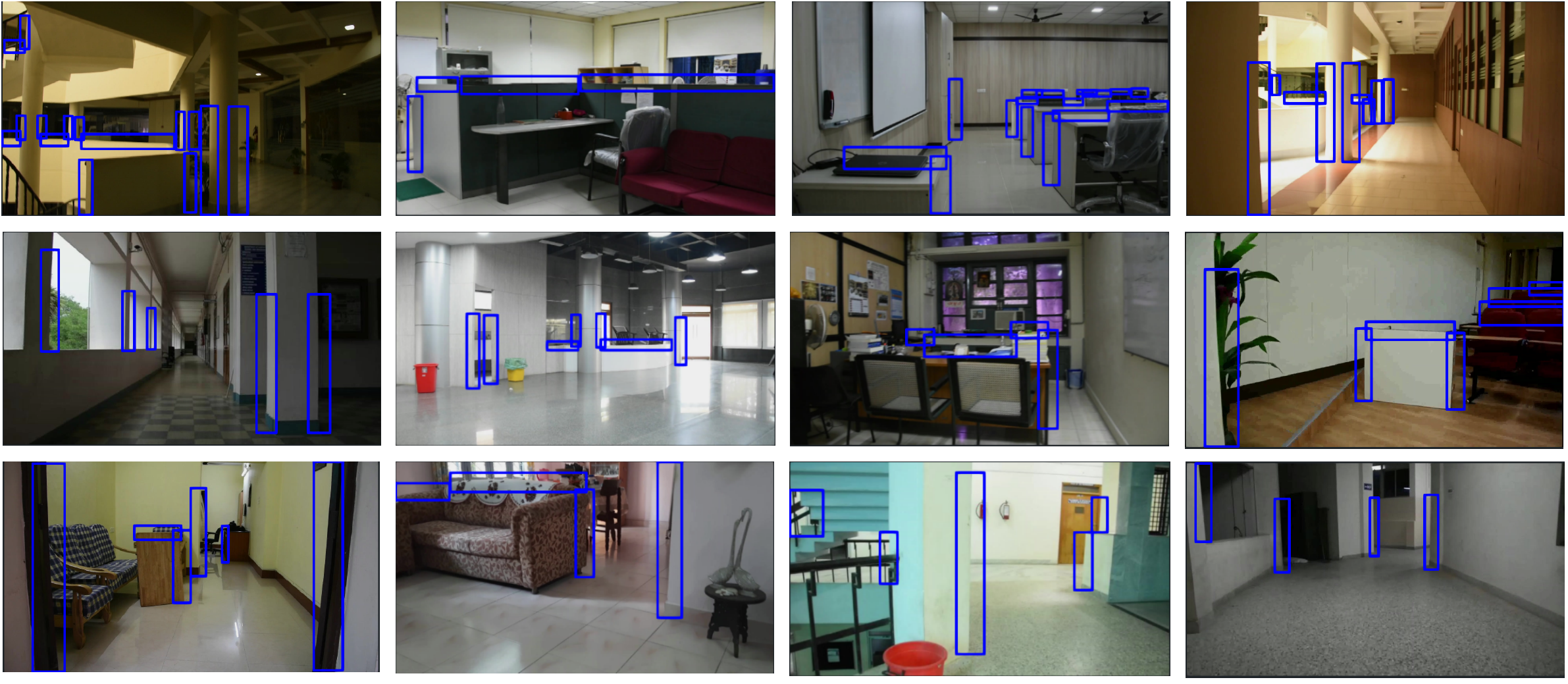}
  \caption{Some of the images from our proposed dataset overlaid with ground truth CGL bounding boxes. CGL bounding boxes are indicated in blue. 
  \MYhref[blue]{https://drive.google.com/file/d/1\_hu31vshulC8P2bTL9xouZQRJ0Btqe0u/view?usp=sharing}{Link to the gallery and dataset.} \cite{gallery}}
  \label{gallery}
\end{figure*}

Existing computer vision datasets do not contain CGL centric real-world images, so we have curated a novel collection of images for CGL detection. Images of our dataset were captured using a Nikon D7200 DSLR camera. We intentionally avoided any humans in the view, to make the images of our dataset realistic from the point of view of the application (counter-insurgency operations) and to avoid violation of ethical norms. Our dataset consists of 1400 real-world images depicting diverse environments. To build robust models that can handle varied lighting conditions, we have captured some of the images in broad daylight and some of the images during nighttime in normal lighting conditions. As will be the case with images encountered by the model in the real-world, images in our dataset contain naturally occurring shadows and glares. Images were captured in corridors, residential houses, offices, labs, classrooms, large dining halls, seminar halls, etc. The majority of our images (sample images are shown in figure \ref{gallery}) depict indoor scenes and a few images depict building premises. Our images are of dimension 1080 $\times$ 1920 (H $\times$ W). 
\\
We experimented with two train/test splits: split 1 and split 2, each containing non-overlapping sets of image samples in train vs test partitions. In split 1, the images in the test set (offices, lobbies, etc) are chosen from scene conditions and locations different from those in the train set (classrooms, houses, etc). This ensures that the locations are unknown for the model at test time. In split 2, few images ($\approx$17\%) in the test set are also from those conditions/locations appearing in the train set too. Split 1 is more challenging as compared to split 2, because the model trained on the training set of split 1 would not have seen test set locations and thus the model would have to be robust enough to not overfit to any inconspicuous biases present in the dataset (train set), to generalize well on new locations. Both the training sets contain around 80\%   of the images and the test sets contain the rest of the images ($\approx$20\%).

\subsection{Annotations}
Annotation involved drawing bounding boxes around CGLs as in the case of object detection \cite{lin2014microsoft,Everingham15}. For annotating our dataset, we used an online interactive annotation tool named Supervisely. Annotations were performed in two phases. In the first phase, bounding boxes were drawn around CGLs and in the second phase, every image was verified by at least three other annotators followed by a master annotator, and minor corrections/adjustments were made to the bounding boxes wherever required. 

To give a physical analogy from the real world:  imagine the scenario of a hide-and-seek game played indoors by children. As a child (player) enters a room and looks for potential places for hideout, he/she must identify zones around all such objects, each of which has the potential to obscure another object or a friend. All edges of an object are not important to explore. The child is knowledgeable enough with the experience to seek only noteworthy places to explore, based on scene analysis by the human brain. This scenario was described and explained to annotators at the beginning of their work, with examples. So the job of the annotators mainly involved identification of specific sub-sections of boundaries of obscuring objects, which have either zones for concealment and are adjacent to zones needing further visual exploration for unearthing any occluded object/human.

As discussed earlier, CGL locations in an image are specific boundaries of occluding objects which have the: i) potential for imminent threats, or ii) potential to provide scope for further exploration of the scene. Hence, CGLs in an image usually lie around the boundary/edges of items like furniture, pillars, doors, windows, staircase, packing boxes, etc. Image regions around specific edges of these items are treated as CGLs if they have the potential (3D space) to occlude and accommodate an object or person. Anyone (or any item) hiding behind any obscuring item may appear in the view from these CGL regions, making these regions locations of threat as well as of interest for further exploration of the environment.

Although the targets for CGL are specific boundaries of occluding objects, it is hard for both the annotators to delineate them, as well as for detection algorithms to identify such sub-segments of edges and lines. Moreover, threats appearing from behind the occluding objects typically appear from around these boundaries. Hence, we decided to identify CGL as a rectangular template around boundaries of occluding objects, with partial coverage of both the body of the object and its background layer. This made the task of both the annotator and the detection model easier.

 CGL bounding boxes have been annotated such that approximately 40\% of the area inside the bounding box contains the obscuring item and the rest (60\%) consists of background or neighboring objects. This 60-40 ratio is only notional and was given as advice to annotators. CGLs in our dataset are of two types: horizontal (wherein the height of the bounding box enclosing the CGL is less as compared to its width) and vertical (wherein the width of the bounding box is less as compared to its height). 

For the height of vertical CGLs, the average human’s height (as qualitatively perceived in the scene by an annotator) was used as a measure of maximum range to annotate. The width of a vertical CGL and the height of a horizontal CGL depends on the size of obscuring object, but we have tried to follow the convention of keeping these quantities close to 100 pixels. Our dataset contains $\approx$15K CGLs, out of which $\approx$59\% are vertical CGLs and $\approx$41\% are horizontal CGLs. Annotations are stored in MS COCO JSON format [30].

CGL detection model should be able to detect CGLs irrespective of the obscuring item. So, to support the training of deep models that can learn the underlying characteristics of CGL without overfitting to the obscuring items present in our CGL detection dataset, we have tried to incorporate a diverse set of obscuring items such as door, staircase, pillar, sofa, window, chairs, etc. Although furniture objects, pillars, and obstructions constitute the majority of the obscuring items in our dataset, their appearances do vary significantly across environments.

\section{Adaptation of existing models for CGL Detection}
As CGL detection involves the identification and localization of RoIs, we can either adapt models designed for object detection or we can use binary semantic segmentation by considering pixels inside CGL bounding boxes to be of one class (CGL) and the rest of the image to be of another class (non-CGL/background). In this section, we describe the object detection and segmentation models we have adapted and trained (from scratch) for CGL detection as part of our initial exploration.

\subsection{YOLOv3}
As YOLOv3 \cite{farhadi2018yolov3} is one of the standard models for most detection tasks, we adapt and train it for CGL detection. However, since CGL detection is much more challenging when compared with object detection, we observe that YOLOv3 (proposed for object detection) fails to capture the complex relationship between CGLs and the neighboring obscuring items.

To classify any region of an input image as a CGL, spatial context is as important as the features of the concerned region. However, existing object detection models do not take into account the spatial context for the classification of RoIs. On the other hand, segmentation models can take into account the spatial context. Thus we map CGL detection to a segmentation task and explore several existing segmentation models for CGL detection.

\subsection{MobileNetv2}
CGL detection could find direct application in the field of robotics. Hence, we consider MobileNetv2 \cite{howard2017mobilenets,sandler2018mobilenetv2} for CGL segmentation. MobileNetv2 provides an excellent option if the model is to be deployed on a low-end device, as it uses depthwise separable convolutions to reduce the number of model parameters. It uses an inverted residual block where skip connections are used to connect linear bottleneck layers.

\subsection{HRNetv2}
Most of the CNN-based feature extractors gradually reduce the resolution of the feature map, but HRNetv2 \cite{sun2019high} maintains high-resolution representation throughout the process. It employs parallel high-to-low resolution subnetworks, which exchange information at each layer to produce good high-resolution representation. HRNetv2 has been shown to perform very well on segmentation tasks. Hence, we incorporate it for CGL segmentation.

\begin{figure*}[ht]
  \centering
  \includegraphics[scale=0.172]{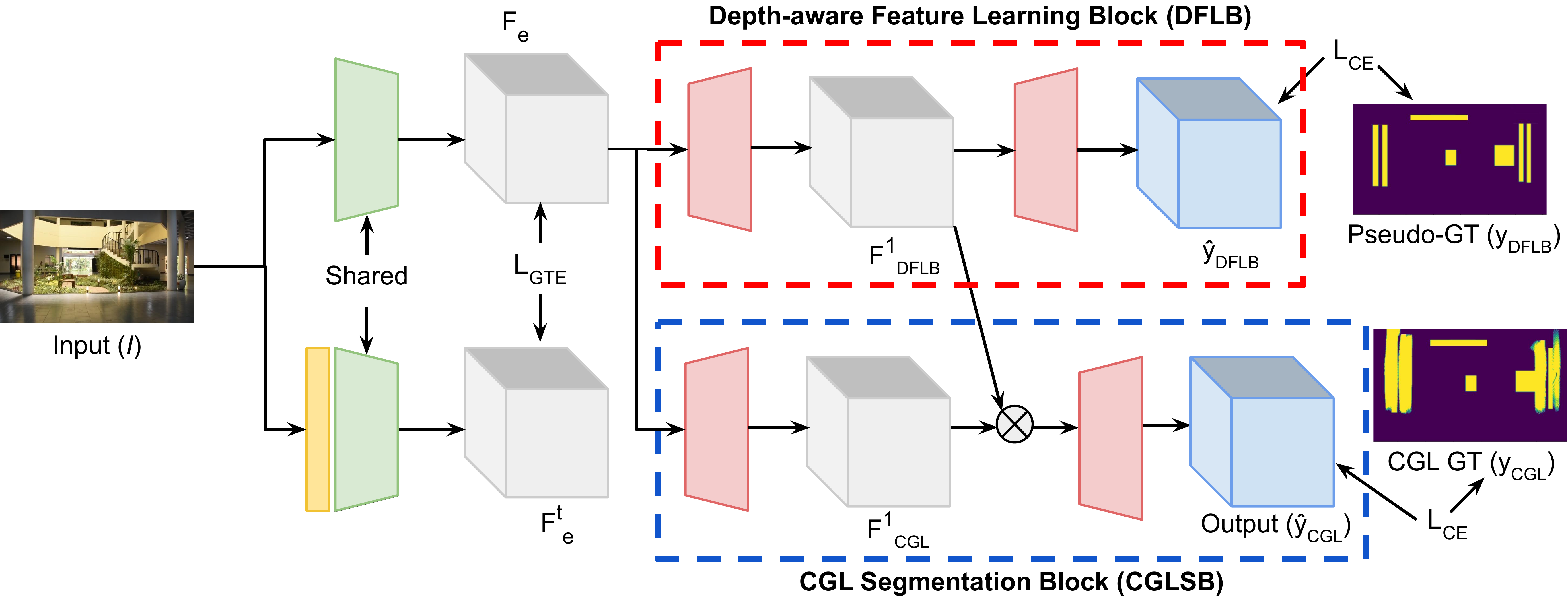}
  \includegraphics[scale=0.202]{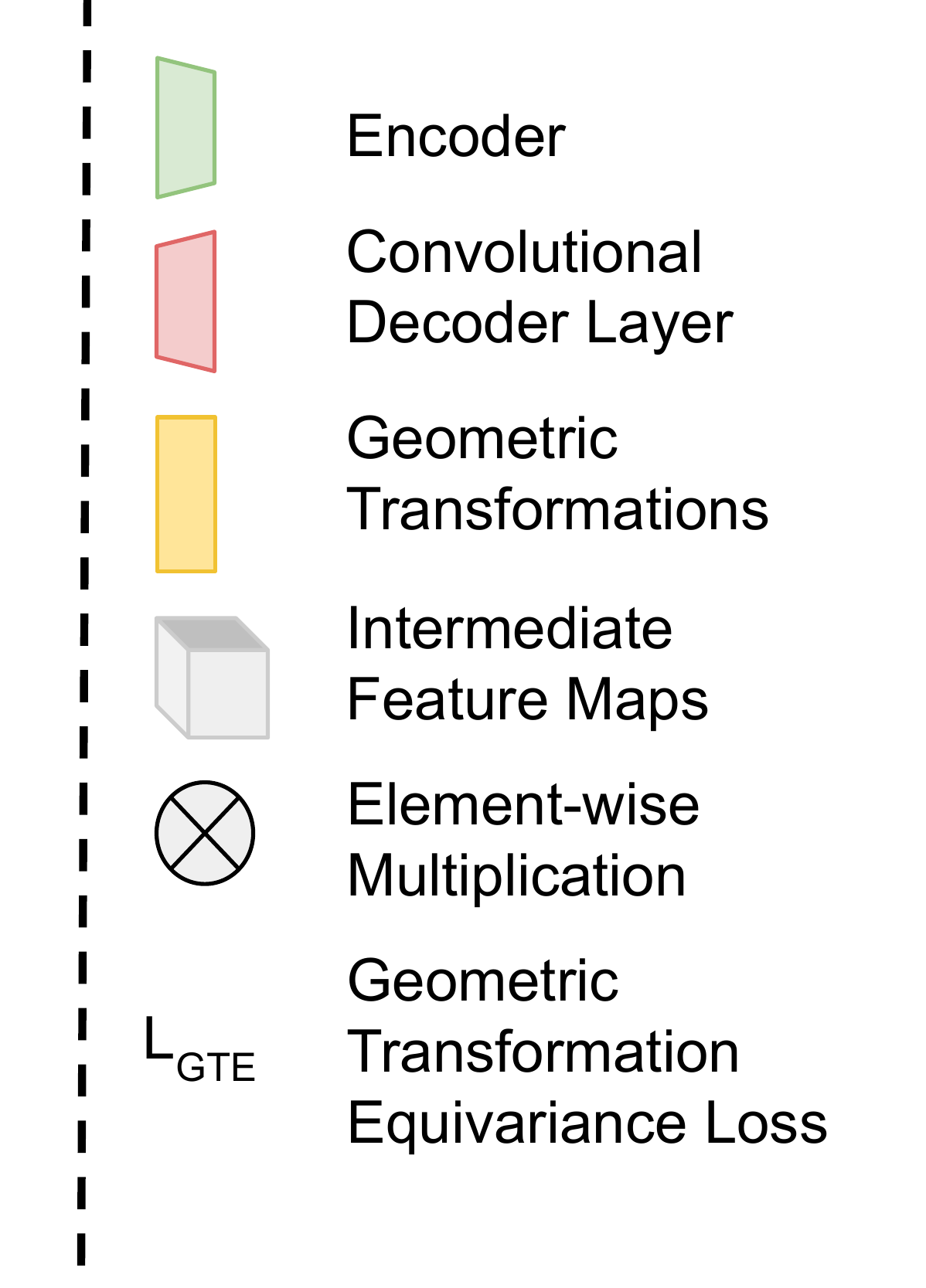}
  \caption{The proposed architecture for CGL detection using binary semantic segmentation. The dashed red box (top) encloses the proposed Depth-aware Feature Learning Block (DFLB), which aids in learning depth-aware features. The dashed cyan box (bottom) encloses the CGL segmentation block (CGLSB).}
  \label{architecture_final}
\end{figure*}

\section{Proposed Method}
\begin{figure*}[ht]
  \centering
  \includegraphics[scale=0.32]{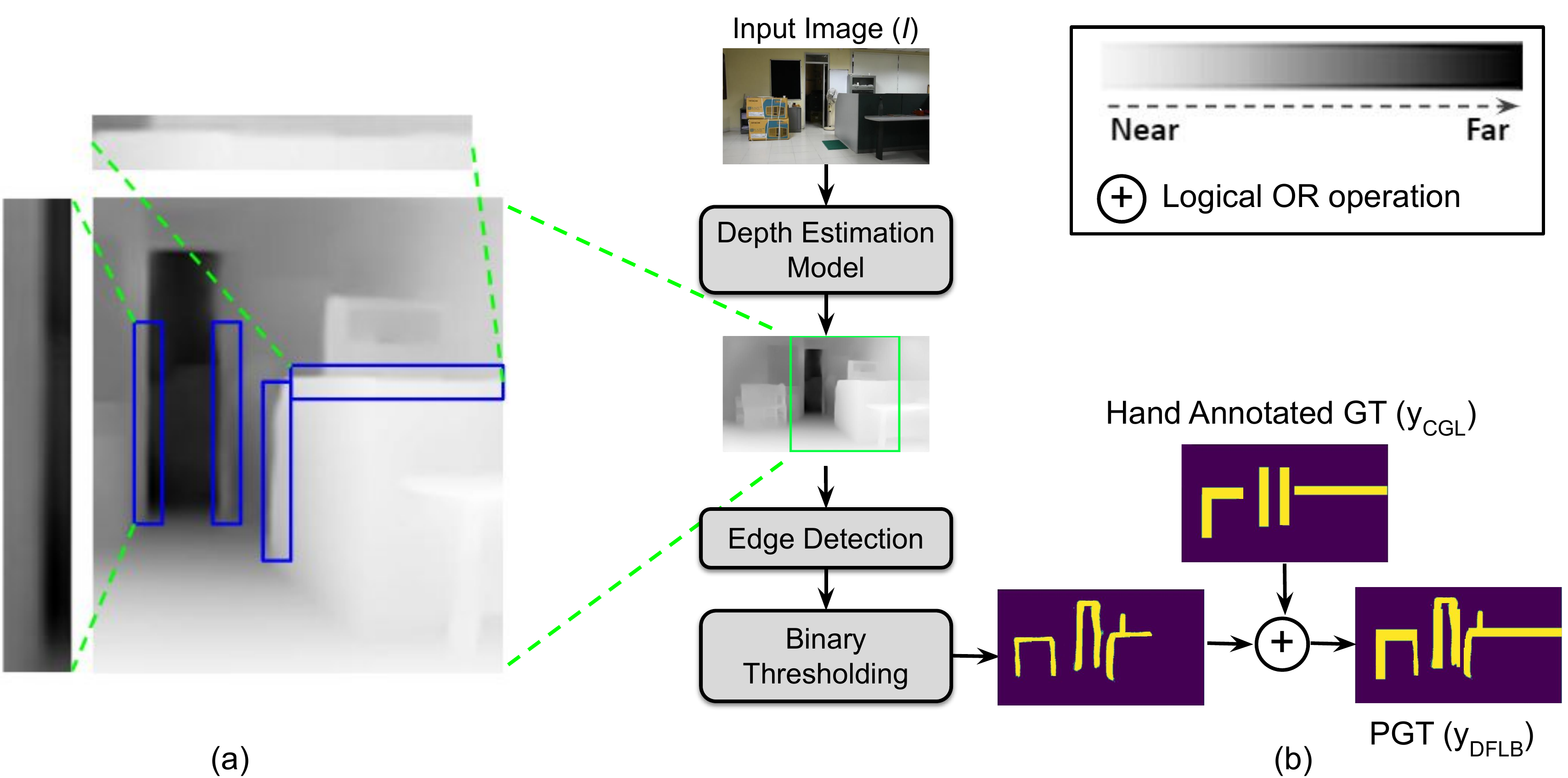}
  \caption{(a) shows a part of depth map for the input image shown in (b). GT CGL bounding boxes have been overlaid on the depth map and some of the bounding boxes have been enlarged for better visualization of the depth pattern in a CGL. This depth pattern will henceforth be referred to as the "CGL depth pattern". (b) shows the flowchart depicting the process of generation of Pseudo-GT (PGT). The green box over the depth map indicates the area that has been enlarged and shown in (a), to clearly exhibit some samples of "CGL depth pattern". Yellow blobs in PGT represent regions having "CGL depth pattern".}
  \label{p_gt_final}
\end{figure*}

We propose a segmentation-based model for CGL detection (shown in figure \ref{architecture_final}). Most RGB segmentation models contain a single encoder and a single decoder. For our method, we propose an additional auxiliary decoder called Depth-aware Feature Learning Block (DFLB), which facilitates the joint learning of RGB and depth features by the common RGB encoder and also guides the CGL Segmentation Block (CGLSB) in attending to regions having depth pattern suitable for CGL detection. Illustration of CGL depth pattern is shown in figure \ref{p_gt_final} (a). Although our model resembles multi-task learning networks \cite{Misra_2016_CVPR,caruana1997multitask}, our main novelty lies in learning relevant depth features using multi-task learning setting. We also propose to use two novel feature-level loss functions for the self-supervision of the encoder (feature extractor). These loss functions are described in detail in sections 5.5 and 5.6.

\subsection{Notations}
Given an RGB input image $I$ $\in$  $R^{3\;\times\;H\;\times\;W}$, the goal of semantic segmentation is to generate a pixel-level classification map with size $H \times W$, where each pixel is assigned to one of the K categories. Number of categories (K) is 2 (CGL and non-CGL/background) in case of CGL segmentation. Input image is first passed through the common encoder to get encoder feature map denoted by $F_{e}$, where $F_{e}$ $\in$ $R^{d\;\times\;H/4\;\times\;W/4}$. Encoder feature maps are passed to both DFLB and CGLSB that generate label maps $\hat{y}_{\DFLB}$ and $\hat{y}_{\CGL}$ respectively as output, where, $\hat{y}_{\DFLB}$, $\hat{y}_{\CGL}$ $\in$ $R^{2\;\times\;H/4\;\times\;W/4}$. We denote the ground truth segmentation masks for DFLB and CGLSB by $y_{\DFLB}$ and $y_{\CGL}$ respectively, where $y_{\DFLB}$,$y_{\CGL}$ $\in$ $R^{H/4\;\times\;W/4}$ (binary maps). Element-wise fusion is performed between the output of first layer of DFLB and that of CGL segmentation block. These intermediate outputs are represented by $F^{1}_{\DFLB}$ and $F^{1}_{\CGL}$ $\in$  $R^{d/4\;\times\;H/4\;\times\;W/4}$ respectively.
\subsection{Encoder}
To extract rich high-resolution features from the RGB input image, we use HRNetv2 as the encoder. However, in the quantitative results section, we show that our method works well irrespective of the encoder used. Encoder feature maps $F_{e}$ are passed to both the decoder blocks, DFLB and CGLSB. The loss components from DFLB and the CGLSB together optimize the encoder weights such that the encoder feature maps capture necessary information for satisfying the objectives of both the decoder blocks simultaneously.

\subsection{Depth-aware Feature Learning Block (DFLB) - Auxiliary decoder}
Although the presence of CGL depth pattern alone does not make any region a CGL, we can infer from figure \ref{p_gt_final} (a) that, depth information provides an excellent cue for CGL detection, since there is an abrupt change in depth values across the edges of obscuring items contained in CGLs. Thus a good encoder should extract features that contain relevant depth information.

To achieve this, we employ an additional auxiliary decoder branch which is trained to perform CGL depth pattern segmentation. As the name suggests, CGL depth pattern segmentation involves pixel-level segmentation of the input image into two classes (One of the classes represents pixels constituting a CGL depth pattern and the other class represents the rest of the pixels). As there is a high correlation between depth information and the desired output for CGL depth pattern segmentation, the segmentation loss for DFLB would compel the common encoder to extract depth-aware features. Hence the additional decoder has been named Depth-aware Feature Learning Block (DFLB).

The first layer of DFLB is composed of one 2D convolutional layer followed by batch normalization and ReLU. For supervision of DFLB, we generate pseudo ground truth (PGT) as illustrated in figure \ref{p_gt_final} (b). PGT ($y_{_{DFLB}}$) is a binary map representing regions having depth pattern similar to that in a CGL by 1s and other regions of the image by 0s. PGT generation involves the following sequence of steps: 1) Pass RGB images to any depth estimation model \cite{hu2019revisiting} to get the depth maps ($I_{depth}$). 2) Smoothen the depth maps. 3) Apply Sobel filter (Canny edge detector does not give superior results) to the smoothened depth map to obtain regions having CGL depth pattern. 4) Squash the gradient magnitudes using the sigmoid function.
\begin{equation}
  g = \phi (G * (K_{avg} * I_{depth})) \label{eq1}
\end{equation}
where, * represents the 2D convolutional operation, $\phi$ is sigmoid function, 
\[ G = 
\begin{bmatrix}
-1 & 0 & 1\\
-2 & 0 & 2\\
-1 & 0 & 1\\
\end{bmatrix} 
\]
and the smoothing kernel $K_{avg}$ is a 11x11 averaging kernel.\\
5) Apply binary thresholding to the output of Step 4.
\begin{equation}
    b = 
\begin{cases}
    1,& \text{if } g\geq Th \text{ (where, Th is a gradient threshold)} \\
    0,              & \text{otherwise} \label{eq2}
\end{cases}
\end{equation}
6) Combine the output obtained in step 5 with the hand-annotated GT ($y_{_{CGL}}$) using logical OR operation ($\xor$), to retain all GT CGL locations in PGT ($y_{_{DFLB}}$), as:
\begin{equation}
  y_{\DFLB} = b \; \xor \; y_{\CGL} \label{eq3}
\end{equation}
\\
\noindent
The loss function used to train DFLB is given as,
\begin{equation}
  \mathcal{L}_{\DFLB} = \mathcal{L}_{_{CE}}(\hat{y}_{\DFLB},y_{\DFLB}) \label{eq4}
\end{equation}
where, $\mathcal{L}_{_{CE}}$ represents cross-entropy loss.

\subsection{CGL Segmentation Block (CGLSB)}
Part of the proposed architecture (shown in figure \ref{architecture_final}) marked using a dashed cyan box indicates CGLSB. This block is responsible for performing CGL segmentation, with encoder feature maps ($F_{e}$) and output of the first layer of DFLB ($F_{DFLB}^{1}$) as input. CGLSB comprises the same sub-layers as DFLB but the weights are different for the two blocks, and they are supervised using different ground truth masks. CGLSB is supervised using hand-annotated CGL ground truths encoded as binary segmentation masks ($y_{\CGL}$). In $y_{\CGL}$, pixels inside GT CGL bounding boxes are represented by 1s and other image pixels are represented by 0s. Output of the first layer of DFLB ($F_{DFLB}^{1}$) contains depth information about image regions as it is used for final class score estimation by the second layer of DFLB. It ($F_{DFLB}^{1}$) helps CGLSB to attend to regions containing CGL depth pattern. The output of this block is generated as given below, 
\begin{equation}
  \hat{y}_{\CGL} = f({F^{1}_{DFLB}}\; \circ \; F^{1}_{\CGL}) \label{eq5}
\end{equation}
where, $f$ represents the second layer in CGL segmentation block, which computes the final class scores. $\circ$ represents element-wise multiplication.
\\
\\
Loss function used to train CGL segmentation block is given as: 
\begin{equation}
  \mathcal{L}_{\CGL} = \mathcal{L}_{\CE}(\hat{y}_{\CGL},y_{\CGL}) \label{eq6}
\end{equation}
where, $\mathcal{L}_{\CE}$ again represents cross-entropy loss.

\subsection{Geometric Transformation Equivariance (GTE) Loss}
Any geometric transformations performed on the RGB image should get replicated in the feature space as well, i.e, if features for the input image $I$ are $F_{e}$ and features for the geometrically transformed image are $F^{t}_{e}$, then if we perform the same transformation on $F_{e}$, we should obtain $F^{t}_{e}$. 
In this paper, we have used rotation (by $90^{\circ}$) as the geometric transformation because we want features for horizontally oriented CGLs to be similar to those for vertically oriented CGLs. 
The loss function is given as follows,
\begin{equation}
  \mathcal{L_{_{GTE}}} = \mathcal{L_{_{MSE}}}(F^{t}_{e},F^{T}_{e})  
  \label{eq7}
\end{equation}
where, $F^{T}_{e}$ represents the rotated version of $F_{e}$. Rotation was performed with the channel dimension as the axis of rotation. 

This loss function helps in learning better representations of the input image in a self-supervised manner.
\subsection{Intraclass Variance Reduction (IVR) Loss}
We want the CGL detection model to focus more on the underlying characteristics of CGL (depth pattern and relationship with respect to neighboring regions of the image) and be robust enough to detect CGLs irrespective of whether the obscuring item was seen during training. So to ensure this, we impose a feature-level constraint using the following loss function,
\begin{equation}
  \mathcal{L_{_{IVR}}} = \sum_{i=1}^{K} \frac{1}{D} \sum_{j=1}^{d} var(\hat{y}_{_{CGL}}[i] * F_{e}[j])
  \label{eq8}
\end{equation}
where, K is the total number of classes (2 for CGL segmentation), and d represents the total number of channels in encoder feature map.

This loss function tries to reduce the intraclass variance in feature space and thus forces the model to focus more on (or extract) features that are absolutely necessary for the task at hand. As a result, the CGL detection model starts to rely more on relevant depth features rather than the features describing the obscuring items. Predicted class probabilities are used to get class information for intraclass feature variance estimation.
\\
\\
Total loss for our model is given as,
\begin{equation}
  \mathcal{L} = \alpha * \mathcal{L_{_{CGL}}} + \beta * \mathcal{L_{_{DFLB}}} + \gamma * \mathcal{L_{_{GTE}}} + \delta * \mathcal{L_{_{IVR}}}
  \label{eq9}
\end{equation}
Where, $\alpha$, $\beta$, $\gamma$, $\delta$ are hyperparameters.
\subsection{Testing}
The depth map is not required during testing as it is used just for the supervision of DFLB and the encoder. The output of the model is produced by the CGLSB. DFLB is not used during testing.

\section{Results and Experiments}
\subsection{Evaluation Metric}
We use the standard mean Intersection-over-Union (mIoU) metric for the evaluation and comparison of models. In order to compare object detection models with segmentation models, a common evaluation metric is required. So, we convert the output of object detection model into segmentation masks by assigning pixels inside predicted bounding boxes to CGL class (1s) and the rest of the pixels to non-CGL class (0s).

On an average, CGLs cover about 10\% of the image. So the models usually have a good IoU score for the non-CGL (background) class, which boosts the mIoU score. For this reason, We also report IoU scores for CGL class separately, denoted as CGL IoU.

\subsection{Quantitative Results}
\begin{table}
\centering
  \caption{Performance of existing object detection and segmentation models when used for CGL detection. All models were trained from scratch on the proposed CGL detection dataset. Architecture of the decoder (C1) is same as CGLSB.}
  \begin{center}
  \begin{tabular}{p{3.5cm}|P{1.5cm}P{2cm}P{1.5cm}P{2cm}}
    \toprule
     Model& \multicolumn{2}{c}{Split 1} &\multicolumn{2}{c}{Split 2}\\
    \cline{2-3}\cline{4-5}
     & mIoU & CGL IoU & mIoU & CGL IoU\\
    \midrule
    YOLOv3 \cite{farhadi2018yolov3} & 52.71 & 19.28 & 56.77 & 26.99\\
    MobileNetv2 + C1 \cite{sandler2018mobilenetv2} &  51.24 & 15.03 & 76.72 & 60.42\\
    HRNetv2 + C1 \cite{sun2019high} & 55.31 & 23.36 & 81.95 & 69.75 \\
  \bottomrule
\end{tabular}
\end{center}
\label{results}
\end{table}

\begin{table}
\centering
  \caption{Performance of the proposed CGL segmentation model, with different encoders.}
  \begin{center}
  \begin{tabular}{p{3cm}|P{1.5cm}P{2cm}P{1.5cm}P{2cm}}
    \toprule
     Encoder & \multicolumn{2}{c}{Split 1} &\multicolumn{2}{c}{Split 2}\\
    \cline{2-3}\cline{4-5}
     & mIoU & CGL IoU & mIoU & CGL IoU\\
    \midrule
    MobileNetv2 \cite{sandler2018mobilenetv2} & 54.19 & 20.62 & 78.46 & 63.40\\
    HRNetv2 \cite{sun2019high} & 61.95 & 35.48 & 83.55 & 72.38 \\ 
  \bottomrule
\end{tabular}
\end{center}
\label{results_ours}
\end{table} 
As evident from table \ref{results}, segmentation models are better at CGL detection. Even though MobileNetv2 has significantly less number of parameters, it outperforms YOLOv3 on split 2, supporting our hypothesis that segmentation models have an upper hand at CGL detection. $mAP_{25}$ for YOLOv3 is 9.58 on split 1 and 14.5 on split 2. $mAP_{50}$ is 5.81 and 10.6 respectively. YOLOv5 yields similar results as YOLOv3. Base code for segmentation models was taken from \cite{zhou2018semantic,zhou2017scene}. Our proposed method effectively extracts necessary depth information from only an RGB image as input and gives the best results on both data splits (evident from table \ref{results_ours}). However, there is still a lot of scope for performance improvement, providing evidence for the toughness of our dataset and the hardness of the problem we have attempted to solve. The performance is lower on split 1 for all models, as they are subjected to completely unseen environments at test time (in split 1). In split 2, partial overlap exists between train and test set environments.
\begin{table}
\centering
  \caption{Ablation study using Split 1, showing the effect of each module in the architecture. HRNetv2 was used as the encoder for all these experiments. Table \ref{results} shows the results obtained when all three additional loss components (DFLB, GTE\_Loss, IVR\_Loss) are absent.}
  \begin{center}
  \begin{tabular}{P{2cm}P{2cm}P{2cm}P{2cm}P{2cm}}
    \toprule
     DFLB & GTE\_Loss & IVR\_Loss & mIoU & CGL IoU\\
    \midrule
    \cmark & \xmark & \xmark & 57.76 & 27.64\\
    \xmark & \cmark & \xmark & 57.50 & 26.96\\
    \xmark & \xmark & \cmark & 57.97 & 27.94\\
    \cmark & \cmark & \xmark & 58.06 & 28.05\\
    \cmark & \xmark & \cmark & 58.45 & 28.91\\
    \xmark & \cmark & \cmark & 59.04 & 29.75\\
    \cmark & \cmark & \cmark & \textbf{61.95} & \textbf{35.48}\\ 
  \bottomrule
\end{tabular}
\end{center}
\label{ablation}
\end{table}

Table \ref{ablation} shows how performance varies with the removal of different modules in our architecture. Overall, we observe that the proposed feature-level loss functions complement the supervised segmentation loss, leading to a significant boost in performance.

\subsection{Qualitative Results}
\begin{figure*}[ht]
  \centering
  \includegraphics[scale=0.4]{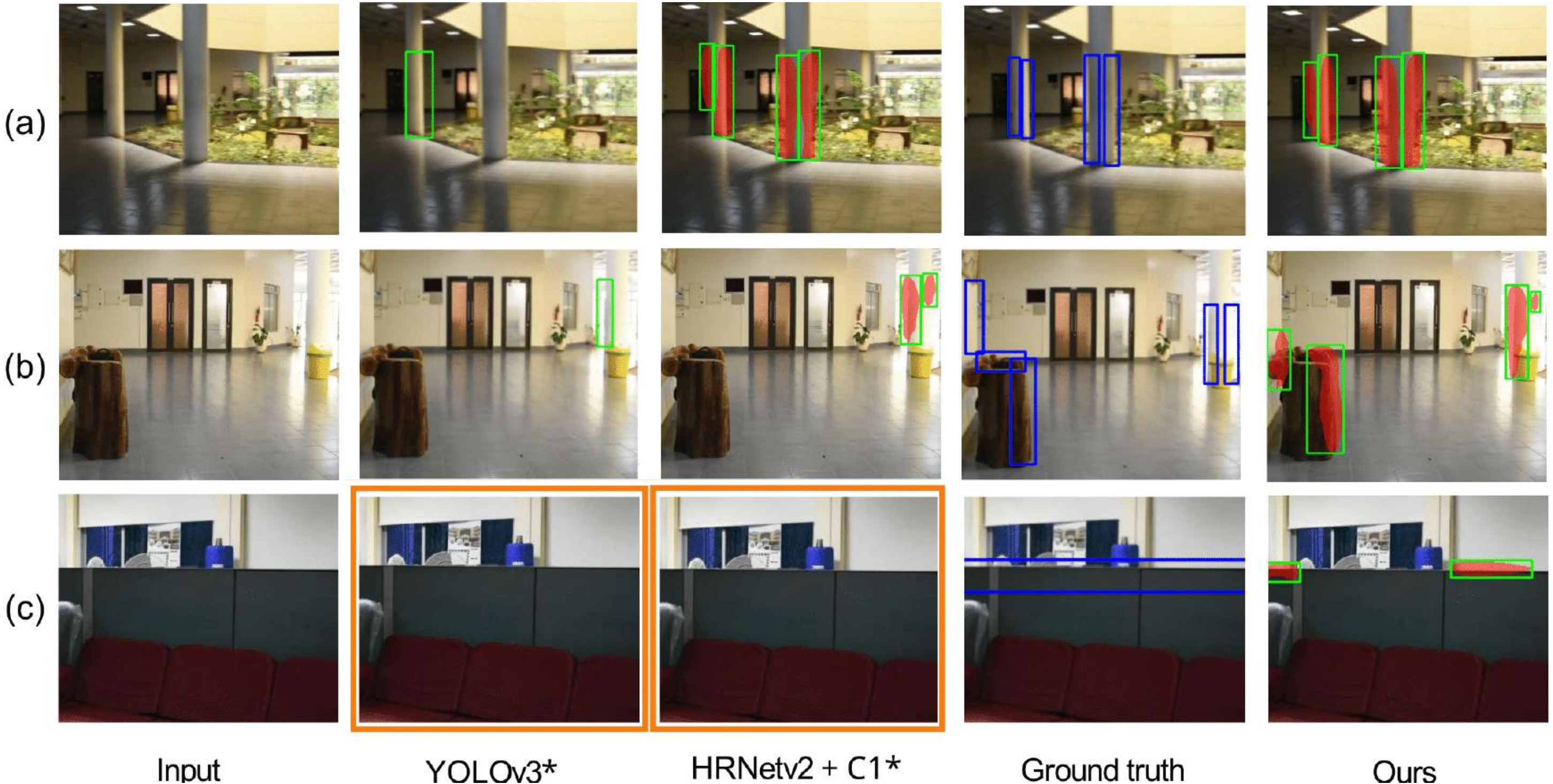}
  \caption{Qualitative results on split 1. To indicate CGL blobs predicted by segmentation models, we have overlaid those regions of the input image with translucent red-colored masks and enclosed them in green bounding boxes. The last column shows the output of our proposed model with HRNetv2 as the encoder. The orange boxes enclose outputs with no detections. In (c), our model detects only certain portions of the CGL. [ * - existing models adapted and trained for CGL detection ]. \MYhref[blue]{https://drive.google.com/drive/folders/13RGqbg4M_78oR59iuWwPCvcYl6Bd_UwL?usp=sharing}{Link to additional qualitative results in the form of videos.} \cite{results}}
  \label{qualitative_results}
\end{figure*}

Figure \ref{qualitative_results} shows qualitative results for existing as well as the proposed method on split 2. It can be seen that existing models fail to capture the depth pattern in CGLs and rely more on edges in the image to detect CGLs. Our proposed model gives more importance to depth information and thus it successfully gets rid of the false detections (see example c) made by baseline models. Additionally, YOLOv3 also struggles to perform accurate localization for CGL detection.

\section{Conclusion and discussion}
In this paper, we propose a new task termed Covert Geo-Location (CGL) Detection. Given an input image, the goal is to identify and localize Covert Geo-Locations (potential hideouts) in the image. We discuss the importance of this task in visual scene understanding and the AI capabilities required to accomplish this task. We provide a CGL detection dataset containing real-world images captured in diverse environments. We demonstrate the importance of depth information for CGL detection and propose a novel segmentation-based Depth-aware Feature Learning Block (DFLB) that facilitates extraction of relevant depth features (from only an RGB image as input) required for the proposed task. We also propose two feature-level loss functions which further complement the supervised loss functions. We provide empirical evidence for the superiority of our model over existing object detection and segmentation models (in the absence of any prior work published in this problem domain). 

Lately, the effective fusion of RGB and depth information has been an active area of research. Although our task is different from existing vision-related tasks and ground truth depth information is not available, existing (or tailored for CGL detection) RGBD segmentation and detection models can also be explored using model-generated pseudo depth map as input along with the RGB image.


%
%
%
\bibliographystyle{splncs04}
\bibliography{samplepaper}

\textbf{ACK: IMPRINT (MHRD/DRDO) GoI, for support}
%




\end{document}